\documentclass[11pt]{article}

\usepackage[margin=1in]{geometry}

\usepackage{mathpazo}

\usepackage{amsmath}
\usepackage{amssymb}
\usepackage{amsthm}
\usepackage[T1]{fontenc}
\usepackage{mathtools}
\usepackage{mdframed}
\usepackage{booktabs}
\usepackage{multirow}
\usepackage{xcolor}
\usepackage{comment}
\usepackage{nicefrac}
\usepackage{mathtools} %

\usepackage{pgfplots}
\pgfplotsset{compat=1.18}
\usepgfplotslibrary{groupplots,colorbrewer}

\usepackage[colorinlistoftodos,textsize=tiny,textwidth=2.1cm,shadow,loadshadowlibrary]{todonotes}

\usepackage[backend=biber, style=numeric, natbib=true,
style=alphabetic,minalphanames=3,maxbibnames=20]{biblatex}
\usepackage{hyperref}
\hypersetup{
  colorlinks=true,
  linkcolor=blue,
  citecolor=blue,
  filecolor=magenta,
  urlcolor=cyan,
}

\newcommand{\method}{\textsc{Magic}}
\newcommand{\trainset}{S} %
\newcommand{\alg}{\mathcal{A}} %
\newcommand{\wvec}{\mathbf{w}}

\newtheorem{theorem}{Theorem}
\newtheorem{example}{Example}

\newtheorem{remark}[theorem]{Remark}

\newcommand{\st}[1]{\mathbf{s}_{#1}}

\newcommand{\replay}{\textsc{Replay}}

\newmdtheoremenv{boxdef}{Definition}

\title{\method{}: Near-Optimal Data Attribution for Deep Learning}
\author{Andrew Ilyas\footnote{Equal contribution; Work done while at MIT} \\
  Stanford Statistics \\ 
  \texttt{ailyas@mit.edu} \and
  Logan Engstrom\footnotemark[1] \\
  MIT EECS \\
  \texttt{engstrom@mit.edu}
}
\date{}

\begin{document}
\maketitle

\begin{abstract}
  The goal of predictive data attribution is to estimate how adding or removing
  a given set of training datapoints will affect model predictions.
  In convex settings, this goal is straightforward (i.e., via the infinitesimal
  jackknife).
  In large-scale (non-convex) settings, however, existing methods are far less
  successful---current methods' estimates often only weakly
  correlate with the ground truth.
  In this work, we present a new data attribution
  method (\method{}) that combines both classical methods and recent advances in
  metadifferentiation~\citep{engstrom2025optimizing} to nearly optimally
  estimate the effect of adding or removing training data on model predictions.
\end{abstract}

\section{Introduction}
A fundamental problem when building machine learning systems is to
predict {\em counterfactuals} about model behavior. For example, scaling laws
\citep{kaplan2020scaling,hashimoto2021model,muennighoff2023scaling}
aim to predict the performance of systems trained with more data and
more compute than
is currently available; interpretability techniques
\citep{kim2018interpretability} predict how models behave under counterfactual
inputs.

Analogously, in this work we study {\em predictive data attribution}
(or {\em datamodeling} \citep{ilyas2022datamodels}),
where the goal is to predict how a model would behave if it
had been trained on a different dataset.
This well-studied problem encompasses, e.g., estimating the effect
(on the resulting trained model's predictions) of modifying a training example
\citep{koh2017understanding}, removing a group of
training examples \citep{koh2019accuracy,bae2022if,park2023trak}, or
adding entire training data sources~\citep{ley2024generalized}.

Predictive data attribution in large-scale settings is challenging: it
requires simulating training a model on a different
dataset without actually training
\citep{guu2023simfluence,ilyas2024data}.
In ``classical'' settings---when learning corresponds to minimizing a
convex loss---statistical tools
like the influence function~\citep{hampel1947influence}
allow us to accurately and
efficiently estimate how different training data choices change trained model
predictions~\citep{rad2018scalable,koh2019accuracy,giordano2019swiss}.
However, in the non-convex settings that are ubiquitous in natural
domains like language/vision, current methods
are less effective. Indeed, the best existing methods produce
estimates that typically (a)
only {\em moderately correlate} with the ground
truth~\citep{basu2021influence,bae2022if,park2023trak} and (b)
incur large absolute error~\citep{bae2022if}.

\subsection{Contributions}
In this work, we make progress towards solving the
well-studied problem of estimating how a model's
predictions would change under different
(counterfactual) deletions of training data
\citep{koh2017understanding,bae2022if,schioppa2022scaling,park2023trak,bae2024training}.
We make two main contributions:

\begin{figure}[ht]
  \centering
  \newcommand{\dataset}{cifar}
  \newcommand{\exampleid}{0}
  \newcommand{\scatteropacity}{0.5}
  \newcommand{\setting}{ResNet-9 on CIFAR-10}
  \newcommand{\fracone}{1}
  \newcommand{\frachtwo}{5}
  \newcommand{\datacsvscatterone}{plots/scatter/scatter_results_cifar/ind_0_dropfrac_0.01.csv}
  \newcommand{\datacsvscattertwo}{plots/scatter/scatter_results_cifar/ind_0_dropfrac_0.05.csv}
  \pgfplotsset{colormap/Set1}

\definecolor{Set1Red}{RGB}{74,124,182}
\definecolor{Set1Blue}{RGB}{142,82,159}
\definecolor{Set1Green}{RGB}{239,134,50}

\usetikzlibrary{calc}

\def\datacsvcorrelations{plots/scatter/correlations_\dataset/dropfrac_\fracone/example_\exampleid.csv}

\def\dataset{wikitext}
\def\datacsvcorrelations{plots/scatter/correlations_\dataset/dropfrac_\frachtwo/example_\exampleid.csv}

\pgfplotstableread[col sep=comma]{\datacsvcorrelations}\datatable

\pgfplotstablegetelem{0}{corr}\of\datatable
\pgfmathprintnumberto[fixed, precision=2]{\pgfplotsretval}{\trakcorrhtwo}

\pgfplotstablegetelem{1}{corr}\of\datatable
\pgfmathprintnumberto[fixed, precision=2]{\pgfplotsretval}{\ekfaccorrhtwo}

\pgfplotstablegetelem{2}{corr}\of\datatable
\pgfmathprintnumberto[fixed, precision=2]{\pgfplotsretval}{\methodcorrhtwo}

\ifdefined\hidelegend
\pgfplotsset{conditional legend style/.style={legend
style={draw=none, fill=none}}}
\else
\pgfplotsset{conditional legend style/.style={
    legend style={
      at={(1.95, 1.45)}, %
      anchor=north east, %
      legend columns=3, %
      /tikz/every even column/.append style={column sep=0.5cm}, %
      inner xsep=5pt, %
      inner ysep=2.5pt,  %
      legend image post style={scale=2.5} %
    }
}}
\fi

\begin{tikzpicture}
  \begin{axis}[
      name=bar_chart, %
      ybar,           %
      width=0.27\textwidth, %
      height=0.3\textwidth, %
      ylabel={Spearman $\rho$},
      xlabel={Method},
      symbolic x coords={A, B, C}, %
      xtick=data,     %
      xticklabels={}, %
      xticklabel style={rotate=45, anchor=east}, %
      nodes near coords, %
      nodes near coords align={vertical},
      nodes near coords style={color=gray}, %
      enlarge x limits=1.5, %
      ymin=0,           %
      ymax=1.2,        %
      axis lines*=left,  %
      bar width=15pt,   %
      title={Correlations}, %
    ]
    \addplot +[fill=Set1Red, draw=none, opacity=0.5] coordinates {(A, 0.234)};
    \addplot +[fill=Set1Blue, draw=none, opacity=0.5] coordinates {(B, 0.347)};
    \addplot +[fill=Set1Green, draw=none] coordinates {(C, 0.957)};
  \end{axis}

  \begin{axis}[
      name=plot1,
      at=(bar_chart.east), %
      anchor=west,         %
      xshift=1.5cm,        %
      width=0.32\textwidth,
      height=0.3\textwidth,
      xlabel={Predicted Loss},
      ylabel={True Loss},
      title={Rand. CIFAR-10 subsets},
      conditional legend style,
      grid=both,
      grid style={line width=.1pt, draw=gray!10},
      major grid style={line width=.2pt,draw=gray!50},
      scaled y ticks=true,
      yticklabel style={/pgf/number format/.cd, fixed, precision=3},
      xticklabel style={/pgf/number format/.cd, fixed, precision=3},
      clip=false,
      after end axis/.code={
        \pgfmathsetmacro{\myxmin}{\pgfkeysvalueof{/pgfplots/xmin}}
        \pgfmathsetmacro{\myxmax}{\pgfkeysvalueof{/pgfplots/xmax}}
        \pgfmathsetmacro{\myymin}{\pgfkeysvalueof{/pgfplots/ymin}}
        \pgfmathsetmacro{\myymax}{\pgfkeysvalueof{/pgfplots/ymax}}
        \pgfmathsetmacro{\linestart}{max(\myxmin, \myymin)}
        \pgfmathsetmacro{\lineend}{min(\myxmax, \myymax)}
        \draw[dashed, gray, very thick, opacity=0.5]
        (axis cs:\linestart,\linestart) --
        (axis cs:\lineend,\lineend);
      }
    ]
    \addplot[
      only marks,
      mark=square*,
      mark size=1.5pt,
      color=Set1Red,
      fill opacity=\scatteropacity,
      draw opacity=\scatteropacity,
      point meta=explicit,
    ] table[x=kron_infl_10, y=true, meta expr=1, col sep=comma]
    {\datacsvscatterone};
    \ifdefined\hidelegend\else
    \addlegendentry{EK-FAC (re-scaled)}
    \fi

    \addplot[
      only marks,
      mark=triangle*,
      mark size=1.5pt,
      color=Set1Blue,
      fill opacity=\scatteropacity,
      draw opacity=\scatteropacity,
      point meta=explicit,
    ] table[x=trak_infl_10, y=true, meta expr=2, col sep=comma]
    {\datacsvscatterone};
    \ifdefined\hidelegend\else
    \addlegendentry{TRAK (re-scaled)}
    \fi

    \addplot[
      only marks,
      mark=*,
      mark size=2pt,
      color=Set1Green,
      fill opacity=1,
      draw opacity=1,
      point meta=explicit,
    ] table[x=infls, y=true, meta expr=0, col sep=comma] {\datacsvscatterone};
    \ifdefined\hidelegend\else
    \addlegendentry{\method{} (unscaled)}
    \fi

    \ifdefined\hidelegend\else
    \addlegendimage{dashed, gray, very thick,
      legend image code/.code={
        \draw[dashed, gray, very thick] (0cm,0cm) -- (0.32cm,0cm); %
      }
    }
    \fi

    \node (annotation_text) at (axis cs: 2.3, 1.9) [align=center,
    text width=4.5cm] {For each color, each dot is a training data subset};
    \coordinate (target_point) at (axis cs: 1.682, 1.87);
    \draw [->, thick, gray, shorten >=1pt] (annotation_text.west) --
    (target_point);

    \node (annotation_text) at (axis cs: 2.32, 1.6) [align=center,
    text width=6.2cm] {\method{} nearly {\em perfectly} predicts loss};
    \coordinate (target_point) at (axis cs: 1.625, 1.6);
    \draw [->, thick, Set1Green, shorten >=1pt]
    (annotation_text.west) -- (target_point);
    \coordinate (target_point_2) at (axis cs: 1.55, 1.54);
    \draw [->, thick, Set1Green, shorten >=1pt]
    (annotation_text.west) -- (target_point_2);
    \coordinate (target_point_3) at (axis cs: 1.827, 1.83);
    \draw [->, thick, Set1Green, shorten >=1pt]
    (annotation_text.west) -- (target_point_3);

    \node (annotation_text) at (axis cs: 2.3, 1.74) [align=center,
    text width=6.2cm] {State-of-the-art attribution methods\\  only
    {\em weakly} correlate with loss};
    \coordinate (target_point) at (axis cs: 1.74, 1.62);
    \draw [->, thick, Set1Red, opacity=0.5, shorten >=1pt]
    (annotation_text.west) -- (target_point);
    \coordinate (target_point_1) at (axis cs: 1.8, 1.75);
    \draw [->, thick, Set1Blue, opacity=0.5, shorten >=1pt]
    (annotation_text.west) -- (target_point_1);
    \coordinate (target_point_3) at (axis cs: 1.683, 1.83);
    \draw [->, thick, Set1Red, opacity=0.5, shorten >=1pt]
    (annotation_text.west) -- (target_point_3);
  \end{axis}
\end{tikzpicture}
  \caption{ \method{} nearly perfectly predicts the effect of training data
    removal. In contrast to the best baselines \citep{park2023trak,grosse2023studying}, 
    \method{} produces estimates
    that both (a) highly correlate with the ground truth effect and (b) are
    well-scaled. {\bf Right:} we plot the predicted loss (from \method{} and 
    the two baselines) against the true loss for a
    randomly chosen test point, each point a training data subset with a random
    1\% of samples removed. For \method{} we plot the predicted loss directly 
    since it is well-scaled; for the baselines, we first rescale the predictions 
    to match the variance of the ground-truth losses.
    {\bf Left:} The average (taken across test examples) 
    Spearman correlation between predicted and true model losses (also known as 
    the LDS \citep{ilyas2022datamodels,park2023trak}, see Section~\ref{sec:single_model_data_attribution}).}
  \label{fig:headline}
\end{figure}
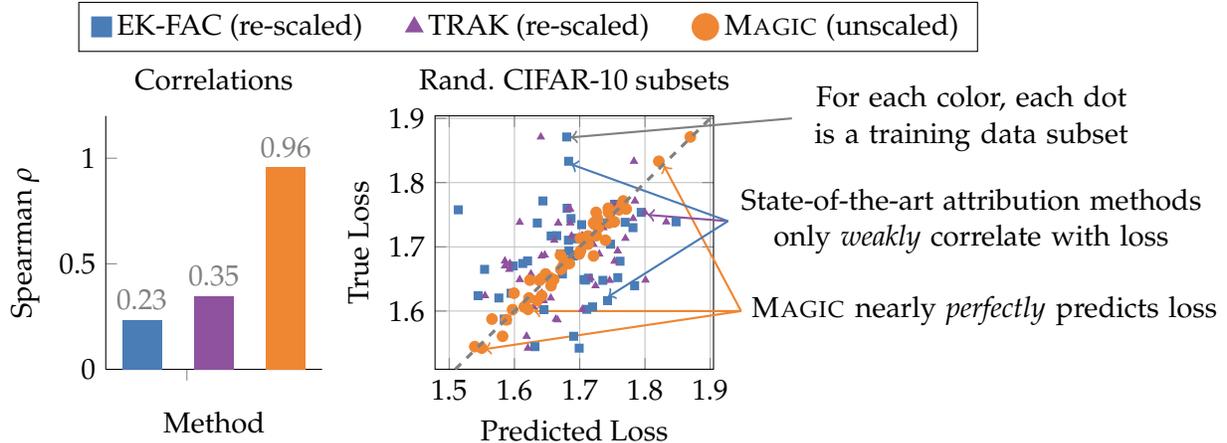

\paragraph{A new perspective (single-model data
attribution).} The inherent randomness of large-scale training makes
attributing specific model predictions to training data
conceptually
challenging \citep{bae2022if,ilyas2022datamodels,nguyen2023bayesian}. After all,
if training on the same data can lead to different models, then
we {\em cannot} predict the variation between these models as a
function of the dataset.
As a result, prior work can only predict how a {\em learning algorithm}
would behave ({\em on average}) if trained on a different dataset,
but not how a {\em specific} model would behave if the training data were
different.

Motivated by this state of affairs,
we introduce a setting called ``single-model'' data attribution.
The goal in this setting is still to predict the behavior of a model under
changes to the training data---the twist is that
we aim to predict how {\em this specific model} would have behaved
under different training data, rather than how a newly initialized
and trained model would have behaved
under different training data.
This subtle change means that: (a) in the single-model setting,
it {\em is} possible to perfectly predict model behavior as a function
of the training data, and (b)
these predictions correspond to how a given ``single model''
would respond to changes in the training data (motivating the name
of the setting), rather than a given learning algorithm.

\paragraph{A new data attribution method (\method{}).}
We present \method{}
(\textbf{M}etagradient-based
  \textbf{A}ttribution via
  \textbf{G}round-truth
  \textbf{I}nfluence
\textbf{C}omputation),
a state-of-the-art
data attribution method.
Our method leverages recent advances in large-scale metagradient calculation
\citep{engstrom2025optimizing} to \textit{exactly} calculate the
influence function \citep{hampel1947influence} for large-scale learning
algorithms.
\method{} accurately estimates how
model predictions respond to random training data deletions---substantially
outperforming existing methods---\textit{even in our more challenging
single-model setting}.
For example (see Figure~\ref{fig:headline}),
\begin{itemize}
  \item When dropping different random 1\% subsets from the training set of a
    ResNet-9 model trained on CIFAR-10, \method{} almost exactly predicts
    ground-truth model losses ($\rho = 0.96$) while existing
    methods' predictions are only weakly correlated ($\rho = 0.25$).
  \item When dropping different random 1\% subsets
    from the training set of a Gemma-2B model trained on instruction tuning
    data, \method{} nearly exactly predicts ground-truth model test losses
    ($\rho = 0.97$) while existing methods perform no better
    than random guessing.
\end{itemize}

\noindent{}Together, the single-model data attribution setting and our new
primitive, \method{}, enable nearly optimally estimating the effect of removing
and adding training data on trained model predictions in modern (deep learning) 
settings, including language modeling and supervised vision tasks.

\section{Data attribution: Notation and problem setup}
\label{sec:problem_setup}
The high-level goal of data attribution is to connect the choice of training
data to model behavior. For example, one may want to use data
attribution to find the training datapoints that cause a given output, or to
surface data that harms accuracy. In this section, we formalize this
goal with the {\em predictive data attribution} (or
\textit{datamodeling}~\citep{ilyas2022datamodels}) framework, which phrases data
attribution as the task of predicting how model behavior changes as a function
of the training data.

Specifically, we view the machine learning pipeline as a three-step process
wherein we
(a) choose training data;
(b) apply a learning algorithm to that data, yielding a trained model;
and then (c) evaluate the trained model.
The goal of predictive data attribution is to construct a function
that {\em directly} predicts the output of step (c) from the choice of
training data in step (a).
To make this more precise (and borrowing from~\citep{ilyas2022datamodels}),
we define the following notation:
\begin{itemize}
  \item Let $\trainset = \{z_i\}_{i=1}^n$ be a pool of $n$ possible training
    examples. We represent {\em datasets} as vectors $\wvec \in \mathbb{R}^n$
    where each entry $w_i$ is an importance weight for the $i$-th example in
    $\trainset$. The importance weight $w_i$ controls the scaling of the loss of
    sample $i$; for example, $w_i=0$ implies that we do not include the $i$-th
    example in the training set, while $w_i > 0$ implies that we do include the
    example (and multiply its loss by $w_i$ during training).
  \item Let $\alg{}: \mathbb{R}^n \to \Theta$ be a {\em learning algorithm}
    mapping datasets---parameterized by importance weight vectors
    $\wvec$---to trained model
    parameters.
    We assume that all aspects of the training setup beyond the
    training data
    are captured by $\alg{}$ (e.g., learning rate, weight decay, etc.).
  \item Let $\phi: \Theta \to \mathbb{R}$ be a {\em measurement
    function} mapping a machine learning model $\theta$ to a scalar measurement
    $\phi(\theta) \in \mathbb{R}$. For example, $\phi(\theta)$ might represent
    the loss of the classifier with parameters $\theta$ on a given test sample.
  \item Let $f: \mathbb{R}^n \to \mathbb{R}$ be the {\em model output function}
    $f$ mapping datasets directly to model outputs, i.e., a composition
    of $\alg{}$ and $\phi$.
\end{itemize}
To illustrate this notation, we instantiate it in the context of linear
regression. In this case, the training pool is a set of $n$ input-label pairs
$\trainset = \{(\mathbf{x}_i \in \mathbb{R}^d, y_i \in \mathbb{R}) \}_{i=1}^n$;
the learning algorithm $\alg{}$ fits a linear model minimizing the average
squared loss {\em weighted} by a given $\wvec \in \mathbb{R}^n$, and the
measurement function $\phi$ evaluates a model's loss on a specific test point
$\mathbf{x}_{test}$, i.e.,
$$\alg{}(\wvec) := \arg\min_\theta \sum_i w_i \cdot (\theta^\top \mathbf{x}_i -
y_i)^2 \qquad \text{and} \qquad \phi(\theta) := (\theta^\top \mathbf{x}_{test} -
y_{test})^2.$$ Then, $f(\wvec) := \phi \circ \alg{}$ maps a data
weighting $\wvec$
to the resulting model's loss on $\mathbf{x}_{test}$.
Now, in this linear regression example, $f(\wvec)$ is easy to directly compute
(in fact, it has a closed form in terms of $\wvec$),
but this is seldom the case.
In general, evaluating $f(\wvec)$ requires re-training a model on the weighting
vector $\wvec$---which can make be very expensive for large-scale models.
This motivates our (informal) definition of predictive data attribution,
given in Definition~\ref{def:predictive_data_attribution} below.

\begin{boxdef}[Predictive data attribution]
  \label{def:predictive_data_attribution}
  \vspace*{-0.5em}
  A \underline{\smash{predictive data attribution}} is an
  \underline{\smash{explicit}} function
  $\hat{f}$ that aims to approximate a model output function $f$.
  In other words, for a given data weight vector
  $\wvec$, $\hat{f}(\wvec)$ should be fast to compute
  while also accurately predicting $f$, i.e., satisfying
  $\hat{f}(\wvec) \approx f(\wvec)$.
  \vspace*{0.4em}
\end{boxdef}

\subsection{Single-model predictive data attribution}
\label{sec:single_model_data_attribution}
The main goal of this work is to operationalize predictive data attribution
for large-scale (deep) learning algorithms.
A challenge, however, is that these learning algorithms are {\em
non-deterministic},
meaning that the same training dataset can map to \textit{different} model
parameters depending on randomness (e.g., random parameter
initialization or data order shuffling) \citep{zhuang2022randomness,jordan2024variance}.

As a result, when learning algorithms are non-deterministic, we cannot perfectly
predict how choice of data will change model behavior; we can only predict how
it will change the {\em expected} behavior. More precisely, data attribution
methods predict how (on average, over training randomness) a {\em new} model
would behave if retrained from scratch, {\em and not how a specific trained
model would behave if we had changed the training dataset}.

To make this more precise, let $\wvec$ be a weighting vector defining a 
dataset, and let $\hat{f}$ be a data attribution method.
Now, consider the expected difference between our estimator $\hat{f}(\wvec)$,
and the true model output $f(\wvec)$, averaged over training randomness. 
This error decomposes into two~terms:
\begin{equation}
  \label{eq:bias_variance}
  \mathbb{E}\left[(\hat{f}(\wvec) - f(\wvec))^2\right] =
  \underbrace{\left(\hat{f}(\wvec) -
  \mathbb{E}[f(\wvec)]\right)^2}_{\text{Reducible Error}} +
  \underbrace{\mathbb{E}\left[\left(f(\wvec) -
  \mathbb{E}[f(\wvec)]\right)^2\right]}_{\text{Irreducible Error}}
\end{equation}
Looking at each term in \eqref{eq:bias_variance}: the \textit{reducible error}
(or bias) is minimal when $\hat{f}(\wvec) = \mathbb{E}[f(\wvec)]$,
while the irreducible error (or variance) depends only on $f$,
and is constant regardless of the data attribution method $\hat{f}$.
Indeed, the irreducible error
arises from inherent randomness in the model training process, and
thus is fundamentally {\em unattributable} to data. Accordingly, current data
attribution methods can answer questions about algorithms (e.g., {\em ``what
    would happen if we trained a new model on a dataset not containing
    the training
example $x$?''})---but not about individual models.

However, in practice we often want to answer questions about
\textit{individual} models,
not a class of learning algorithms. For example, we might ask a question
like {\em ``what was the effect of training example $x$ on this specific
model?''} 
This question motivates us to define and consider a problem 
that we call {\em single-model data attribution}.

\medskip
\noindent\textbf{Single-model predictive data attribution.}
To understand how choice of data changes \textit{individual} trained models, we
propose a new setting called {\em single-model data attribution}.
Here, we enforce that
the learning algorithm $\alg{}$ and the measurement function $\phi$ are
deterministic (i.e., by fixing data ordering,
parameter initialization, etc.).
This determinism ensures that for any weighting $\wvec$, the model output
$\phi(\alg{}(\wvec))$ is deterministic
(i.e., so that the expected model output that datamodels predict is constant,
and $\text{Var}\left[{\phi(\alg{}(\wvec))}\right] = 0$). In this new setting,
model outputs vary only from changes in training data weights, allowing for
predictive data attribution methods to exactly attribute changes to
data weights. In the language of \eqref{eq:bias_variance}, the irreducible error
is zero.

\begin{remark}[Single-model versus standard predictive data attribution]
  Our motivation for the single-model setting is that in many cases
  we often want to understand the effect of training data on a
  specific model, rather than a class of models.
  Still, even in cases where we \underline{do} care more about 
  the average behavior of a learning algorithm (and not  
  a specific trained model), a near-perfect single-model data attribution
  method can be used to construct a near-perfect standard data
  attribution method by averaging over different learning algorithm seeds.
  We discuss this connection further in Section \ref{sec:discussion}.
\end{remark}

\section{\method{}: Calculating the exact influence function at scale} We now
present \method{}, our method for nearly-optimal single-model data attribution.
The skeleton of our method is conceptually straightforward: we exactly calculate
the influence function in large-scale learning settings.
In this section, we first formally define the influence function;
we then define a specific class of learning algorithms that we will consider;
and finally, we present our method for calculating the exact influence function
for this class of learning algorithms.

\subsection{The influence function} %
\label{sec:influence_fn}
At the core of our method is a statistical primitive
known as the {\em influence function approximation}
\citep{hampel1947influence,koh2017understanding,giordano2019swiss}.
The main idea is to approximate the model output $f$ for a given
data weighting
$\wvec$ using the following first-order Taylor expansion:
\begin{equation}
  \label{eq:infl_fn}
  \hat{f}(\wvec) := f(\mathbf{1}_n) +
  \left(\frac{\partial f(\wvec)}{\partial
  \wvec}\bigg|_{\wvec=\mathbf{1}_n}\right)^\top  (\wvec -
  \mathbf{1}_n);
\end{equation}
The key term in this estimate is the gradient
$\nicefrac{\partial f(\wvec)}{\partial \wvec}$ evaluated at
$\wvec=\mathbf{1}_n$, called the \textit{influence function}.
Intuitively, this term captures the effect of infinitesimally up- or
down-weighting each training example on the model output.
While well-defined, this quantity is not trivial to compute:
after all, it is the gradient ``through'' the process of training
the model with algorithm $\alg{}$.

\paragraph{The main challenge of influence functions in large-scale settings:
computing the gradient.} When the learning algorithm $\alg{}$ is a {\em convex}
optimization algorithm (e.g., linear regression, logistic regression, etc.),
the influence function is straightforward to compute. Indeed, in such settings,
the gradient $\nicefrac{\partial f(\wvec)}{\partial \wvec}$ has a simple closed
form (via implicit differentiation),
and the first-order Taylor expansion \eqref{eq:infl_fn} yields
near-perfect estimates of the model output $f(\wvec)$ will behave on many
choices of new data weightings $\wvec$
\citep{koh2017understanding,rad2018scalable,koh2019accuracy,giordano2019swiss}.

In large-scale, non-convex settings (e.g., in deep learning),
however, the influence function is difficult to compute:
no such closed form exists.
Instead, large-scale data attribution methods that are based on
\eqref{eq:infl_fn}
must {\em approximate} the influence function
\citep{koh2017understanding,bae2022if,park2023trak,bae2024training}.
And while these methods have shown promise,
they are not nearly as effective as the influence function is in
analogous convex
settings.

\subsection{Focus: iterative smooth learning algorithms}
Before describing our procedure, we first formalize the class of
learning algorithms
$\alg{}$ that we will consider, namely ones that are {\em iterative}
and {\em smooth}.
By restricting our focus to this class
of learning algorithms, we ensure that the influence function is well-defined.
Note: this class of algorithms is extremely general and captures,
e.g., large-scale (transformer-based) language model training or deep image
classifier training (sometimes with slight changes from standard
learning algorithms).

\subsubsection{Iterative learning algorithms}
First, we require that the learning algorithm $\alg{}$
is {\em iterative}, i.e., it takes the form
\begin{align}
  \label{eq:iterative_algo}
  \alg(\wvec) := \st{T} \quad \text{for} \quad
  \st{t+1} &:= h_t(\st{t}, \mathbf{g}_t(\st{t}, \wvec))
  \quad \text{ and } \quad \mathbf{g}_t(\st{t}, \wvec) :=
  \sum_{i\in B_t} w_i \cdot \nabla_{\st{t}} \ell(z_i; \st{t}).
\end{align}
Above,
$\st{t}$ is the optimizer state (including model parameters),
which is iteratively updated by a function $h_t$
starting from a fixed initial state $\st{0}$.
We let the number of training steps be $T$;
$B_t \subset [N]$ is a minibatch sampled at step $t$;
and $\ell(z_i; \st{t})$ is the loss on
sample $z_i$ given optimizer state $\st{t}$.

The vast majority of large-scale learning algorithms are iterative
in this sense---we give a few examples below.
Recall that the learning algorithm $\alg{}$ takes as input a
data weighting $\wvec$ over the training set, and outputs
a machine learning model trained on the weighted dataset.
The algorithm thus encapsulates all aspects of the training setup
beyond the training data weights, including the model architecture,
optimizer, and hyperparameters.

\begin{example}[Training an ResNet with SGD]
  Here, the optimizer state $\st{t}$ is the parameter vector
  $\theta_t$ at step $t$,
  the loss function $\ell(z_i; \st{t})$ is the cross-entropy loss of a ResNet
  with parameters $\st{t}$ on a given training example $z_i$,
  and the update function $h_t$ is the SGD update step
  $$h(\st{t}, \mathbf{g}_t) := \st{t} - \eta_t \cdot \mathbf{g}_t,$$
  where $\eta_t$ is the learning rate at step $t$.
\end{example}

\noindent More complex optimizers can be handled by extending the definition
of the optimizer state $\st{t}$ beyond just the parameter vector,
as we show in the following example.
\begin{example}[Training a language model with Adam]
  Here, the optimizer state $\st{t} = (\theta_t, m_t, v_t)$,
  where $\theta_t$ is the parameter vector at step $t$,
  $m_t$ is the first moment estimate of the gradient,
  and $v_t$ is the second moment estimate of the gradient.
  The loss function $\ell(z_i; \st{t})$ is the cross-entropy loss of a language
  model with parameters $\theta_t$ on a given training example $z_i$,
  and the update function $h_t$ is the Adam update step:
  $$h(\st{t}, \mathbf{g}_t) := \left[
    \begin{array}{c}
      \theta_t - \eta_t \cdot \frac{\sqrt{v_t}}{\sqrt{m_t +
        \epsilon_{\mathrm{root}}} +
      \epsilon} \cdot \mathbf{g}_t \\
      \beta_1 \cdot m_t + (1 - \beta_1) \cdot \mathbf{g}_t \\
      \beta_2 \cdot v_t + (1 - \beta_2) \cdot \mathbf{g}_t^2
    \end{array}
  \right]$$
  where $\eta_t$ is the learning rate at step $t$.
\end{example}

\subsubsection{Smooth learning algorithms}
Finally, for predictive data attribution to even be possible (recalling
Definition \ref{def:predictive_data_attribution}), we must ensure that the
learning algorithm $\alg{}$ is (qualitatively) well-behaved as a function of the
data weights $\wvec$.
Indeed, when this is not the case,
it is unlikely that {\em any} simple predictor $\hat{f}$
will be able to accurately predict $f(\wvec)$ from $\wvec$.
To formalize this requirement, we consider learning algorithms $\alg{}$ that are
{\em smooth} in $\wvec$, as described by \citet{engstrom2025optimizing}.
In particular, a learning algorithm $\alg{}$ is smooth in $\wvec$ if,
for any measurement function $\phi$, small perturbations to the data weights
$\wvec$ result in only small changes to the gradient
$\nicefrac{\partial f(\wvec)}{\partial \wvec}$.

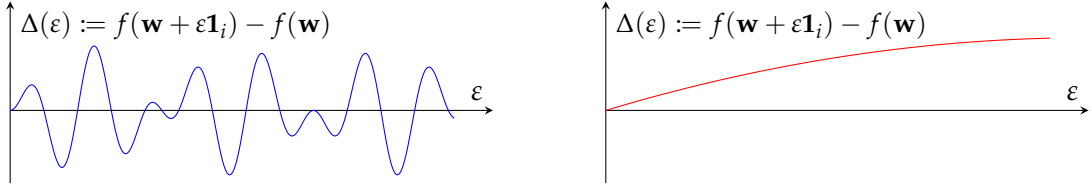
\begin{figure}[h]
  \centering
  \begin{tikzpicture}
  \begin{groupplot}[
      group style={
        group size=2 by 1, %
        horizontal sep=1.5cm %
      },
      width=8cm,       %
      height=4cm,      %
      xmin=0, xmax=0.5,
      ymin=-10, ymax=15,
      axis lines=middle, %
      xtick=\empty,      %
      ytick=\empty,      %
      grid=none,         %
      enlargelimits=false, %
      xlabel={$\varepsilon$}, %
      ylabel={{\small $\Delta(\varepsilon) \coloneqq f(\wvec +
          \varepsilon \mathbf{1}_i) -
      f(\wvec)$}} %
    ]

    \nextgroupplot %
    \addplot[
      domain=0:0.46,
      samples=300, %
      blue %
    ]
    { 9 * sin(deg(90*x)) * sin(deg(20*x)) };

    \nextgroupplot %
    \addplot[
      domain=0:.46,
      samples=100, %
      smooth,      %
      red          %
    ]
    { 40*x - 40*x^2 };

  \end{groupplot}
\end{tikzpicture}
  \caption{Smoothness aids predictive data attribution. We plot the change in
    data weights $\varepsilon$ against the change in model output
    $\Delta(\varepsilon)$ for two hypothetical learning algorithms. On the left
    is a non-smooth setting where the
    gradient $\nicefrac{f(\wvec)}{\wvec}$ varies wildly with
    $\varepsilon$. On the
  right is a smooth setting where the change is well-behaved. }
  \label{fig:smoothness}
\end{figure}

To see why smoothness is necessary to predict $f$ with a simple function, we walk
through a simple example (visualized in Figure \ref{fig:smoothness}).
For a model output function $f$, consider slightly upweighting the
$i$-th training sample,
and measuring the change in the model output $\Delta(\varepsilon) :=
f(\wvec + \varepsilon \mathbf{1}_i) - f(\wvec)$.
If the learning algorithm is smooth, then this change is well-behaved as a
function of $\varepsilon$.
In particular, the change $\Delta(2\varepsilon)$ should be
reasonably approximated by $2\Delta(\varepsilon)$.
On the other hand, if the learning algorithm is \textit{not} smooth,
$\Delta(\varepsilon)$ may change wildly as $\varepsilon$ varies,
precluding any simple prediction method from being able to accurately
predict $f(\wvec)$ from $\wvec$.

\begin{remark}[How restrictive is smoothness?]
  Unlike iterativity, smoothness is not an inherent
  property of standard learning algorithms.
  Indeed, many standard training routines are \underline{not}
  smooth.
  In such cases, there is
  \underline{\smash{no}} good data attribution method $\hat{f}$
  satisfying a natural ``additivity'' property \citep{saunshi2023understanding}
  (ruling~out essentially all known data attribution methods).
  Fortunately, however, as observed by prior work,
  one can often construct a ``smooth counterpart''
  to any given non-smooth learning algorithm \citep{engstrom2025optimizing}.
  Motivated by this finding (and the impossibility above),
  we focus our attention on smooth learning algorithms.
\end{remark}

\subsection{Calculating the exact influence function}
To compute the exact influence function for a model output function $f$
that is the output of an iterative smooth learning algorithm $\alg{}$,
we leverage recent developments in {\em metagradient} calculation
\citep{maclaurin2015gradient,franceschi2017forward,lorraine2020optimizing,engstrom2025optimizing}.
A metagradient is a gradient of a machine learning model's output
with respect to a design choice made prior to training.
In recent work, \citet{engstrom2025optimizing} present an algorithm
called \replay{}
that {\em exactly} calculates the metagradient for iterative and smooth
learning algorithms.

\begin{figure}[t]
  \centering
  \begin{tikzpicture}[
    node distance=2cm,
    state/.style={rectangle, draw, minimum size=0.8cm},
    func/.style={rectangle, rounded corners, minimum size=0.6cm},
    arrow/.style={->, >=stealth, thick}
]

\node[state] (s0) {$\st{0}$};
\node[state, right of=s0] (s1) {$\st{1}$};
\node[state, right of=s1] (s2) {$\st{2}$};
\node[right of=s2] (dots) {$\cdots$};
\node[state, right of=dots] (sT-1) {$\st{T-1}$};
\node[state, right of=sT-1] (sT) {$\st{T}$};
\node[func, right of=sT] (phi) {$\phi(\st{T})$};

\draw[arrow] (s0) -- (s1);
\draw[arrow] (s1) -- (s2);
\draw[arrow] (s2) -- (dots);
\draw[arrow] (dots) -- (sT-1);
\draw[arrow] (sT-1) -- (sT);
\draw[arrow] (sT) -- (phi);

\coordinate (bottom) at (0,-1.7);
\node[func] at ($(s0)!0.5!(s1) + (0,-1.7)$) (g0) {$\mathbf{g}_0$};
\node[func] at ($(s1)!0.5!(s2) + (0,-1.7)$) (g1) {$\mathbf{g}_1$};
\node at ($(s2)!0.5!(dots) + (0,-1.7)$) (gdots) {$\cdots$};
\node[func] at ($(dots)!0.5!(sT-1) + (0,-1.7)$) (gT-2) {$\mathbf{g}_{T-2}$};
\node[func] at ($(sT-1)!0.5!(sT) + (0,-1.7)$) (gT-1) {$\mathbf{g}_{T-1}$};

\draw[arrow] (s0) -- (g0);
\draw[arrow] (g0) -- (s1);
\draw[arrow] (s1) -- (g1);
\draw[arrow] (g1) -- (s2);
\draw[arrow] (sT-1) -- (gT-1);
\draw[arrow] (gT-1) -- (sT);

\draw[arrow] (s2) to[out=-45,in=135] (gdots);
\draw[arrow] (dots) to[out=-45,in=135] (gT-2);
\draw[arrow] (gT-2) -- (sT-1);

\node[func] at ($(g0) + (0,-1.5)$) (w0) {$\wvec$};
\node[func] at ($(g1) + (0,-1.5)$) (w1) {$\wvec$};
\node at ($(gdots) + (0,-1.5)$) (wdots) {$\cdots$};
\node[func] at ($(gT-2) + (0,-1.5)$) (wT-2) {$\wvec$};
\node[func] at ($(gT-1) + (0,-1.5)$) (wT-1) {$\wvec$};

\draw[arrow] (w0) -- (g0);
\draw[arrow] (w1) -- (g1);
\draw[arrow] (wT-2) -- (gT-2);
\draw[arrow] (wT-1) -- (gT-1);

\end{tikzpicture}
  \caption{Forward computation graph for a model output function $f$
    mapping from data weights $\wvec$ to the model output.
    The exact influence function $\nicefrac{\partial f(\wvec)}{\partial \wvec}$
    is the {\em metagradient} of the model output with respect to the
  data weights $\wvec$.}
  \label{fig:computation_graph}
\end{figure}
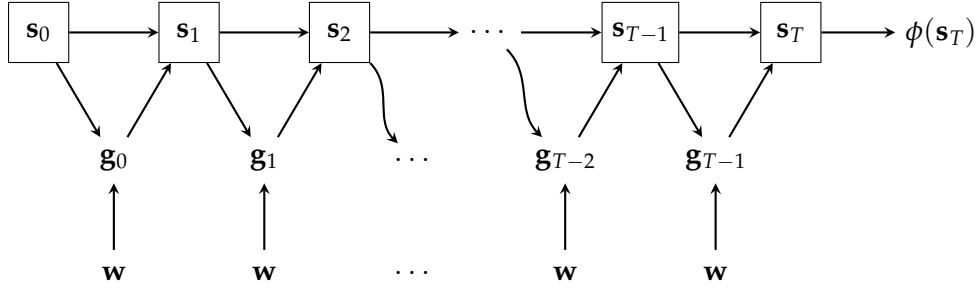

\paragraph{Calculating the influence function with \replay{}.}
Observe that when the design choice is the data weighting $\wvec$,
the metagradient---the gradient of the model output $f$ with respect to
$\wvec$---is precisely the influence function.
We can thus directly apply the \replay{} algorithm
\citep{engstrom2025optimizing}
to calculate the exact
influence function.
Adapted to our setting, \replay{} calculates the metagradient by
exploiting the following identity, which follows from the chain rule
applied to the computation graph illustrated in Figure
\ref{fig:computation_graph}:
\begin{align}
  \nabla_{\wvec} f(\wvec)
  &= \sum_{t=0}^{T-1} \underbrace{
    \frac{\partial f(\wvec)}{\partial \st{t+1}} \cdot
    \frac{\partial h_t(\st{t}, \mathbf{g}_t(\st{t}, \wvec))}{\partial \wvec}
  }_{\text{contribution of $\wvec$ to $f(\wvec)$ through $\st{t+1}$}}
  \label{eq:replay_meta_grad}
\end{align}

Motivated by this observation, \replay{} operates as follows:
\begin{enumerate}
  \item Initialize $\Delta_T = \frac{\partial f(\wvec)}{\partial
    \st{T}} = \nabla_{\st{T}} \phi(\st{T})$
  \item For each $t = T-1, \ldots, 0$:
    \begin{enumerate}
      \item Load state $\st{t}$ and minibatch $B_t$
      \item Calculate $\beta_t = \nabla_{\wvec} \left(h_t(\st{t},
        \mathbf{g}_t(\st{t}, \wvec))^\top \Delta_{t+1}\right)$,
        the contribution of $\wvec$ to $\phi(\st{T})$ through the
        $t$-th step of the learning algorithm
      \item Advance $\Delta_t = \nabla_{\st{t}} \left(h_t(\st{t},
        \mathbf{g}_t(\st{t}, \wvec))^\top \Delta_{t+1}\right)$,
        which is $\nicefrac{\partial f(\wvec)}{\partial \st{t}}$
    \end{enumerate}
  \item Return $\beta = \sum_{t=0}^{T-1} \beta_t$ as the exact
    influence function
\end{enumerate}

\noindent By leveraging an efficient data structure to
load the states and minibatches,
\replay{} is able to calculate the exact influence function
for IDS model output functions at a computational cost of
$T + T\log(T)$ total training steps, and $\log(T)$ memory.
We refer the reader to \citet{engstrom2025optimizing} for a complete
description of the algorithm.

\section{Evaluation}
\label{sec:results}
In this section we evaluate \method{} across a number of domains.
In particular, we compare \method{} with two of the most successful
recent data attribution techniques: TRAK \citep{park2023trak} and
EK-FAC \citep{grosse2023studying}; see Appendix~\ref{app:baselines} for the
specifics of these baselines.
Across the board, we find that \method{} provides near-perfect
predictions of how model outputs change when we drop data
(at random) from the training set.

\paragraph{Evaluation metric.}
Recall (from Section~\ref{sec:problem_setup}) that the goal of predictive
data attribution is to predict how a model's output changes as a function of
the model's training data. In order to evaluate the quality of these
predictions,
we adopt the {\em linear datamodeling score} (LDS)
\citep{ilyas2022datamodels,park2023trak}
as our evaluation metric. To compute LDS for a given model output function
$f$ and corresponding data attribution method $\hat{f}$, we follow
the following steps:
\begin{enumerate}
  \item Sample $n$ fixed-sized subsets of the training set,
    which we represent as binary data weights
    $\wvec^{(1)}, \ldots, \wvec^{(n)} \in \{0,1\}^N$,
    where $N$ is the number of total training samples.
    Given a drop-out fraction $p \in [0,1]$, we sample each data weight
    vector $\wvec^{(i)}$ by dropping $pN$ random training samples
    from the training set.

  \item For each data weight vector $\wvec_i$:
    \begin{enumerate}
      \item Compute the {\em true} model output function
        $f(\wvec_i)$ by training a model on the training set with
        data weights $\wvec_i$
        and evaluating the measurement of interest on the trained model.
      \item Compute the {\em predicted} model output function
        $\hat{f}(\wvec_i)$ via the data attribution method $\hat{f}$.
    \end{enumerate}

  \item We compute the LDS as the Spearman correlation between the predicted
    output and the true output over all $n$ data weight vectors, i.e.,
    \begin{align}
      \label{eq:lds}
      \text{LDS} =  \rho\!\left(\left[\hat{f}(\wvec_i)\right]_{i=1}^n,
      \left[\phantom{\hat{f}}\!\!\! f(\wvec_i)\right]_{i=1}^n\right).
    \end{align}
\end{enumerate}

\paragraph{Settings.}
We study scenarios spanning computer vision and
language modeling.
Each scenario comprises a training dataset $S$,
a learning algorithm $\mathcal{A}$,
and a test set $S_T$. Accordingly, each scenario defines $|S_T|$ data
attribution
tasks, where task $i$ is to predict the loss of $\mathcal{A}$ on the
$i$-th sample
in $S_T$. For each data attribution method we consider,
we compute the {\em average} LDS \eqref{eq:lds} across tasks.
\begin{itemize}
  \item {\bf ResNet CIFAR-10 training}: We train ResNet-9~\citep{jordan202494}
    models on subsets of the CIFAR-10 train set~\citep{krizhevsky2009learning},
    and aim to predict cross-entropy loss on CIFAR-10 test samples.
  \item {\bf GPT-2 Wikitext fine-tuning}: We fine-tune 125M
    GPT-2 models~\citep{radford2019language} on subsets of
    Wikitext~\citep{foundation2022english}, and aim to predict
    (language modeling) loss on 50 different test samples.
  \item {\bf Gemma-2B instruction-tuning}: We fine-tune Gemma-2B
    \citep{team2024gemma} with LoRA \citep{hu2021lora} on 
    subsets of three combined instruction tuning datasets
    (Flan V2, DOLLY, OpenAssistant-1
    \citep{longpre2023flan,conover2023free,kopf2024openassistant}), aiming to predict 
    loss on MMLU \citep{hendrycks2021measuring} samples.
\end{itemize}
See Appendix~\ref{app:expdetails} for the exact details of each scenario.

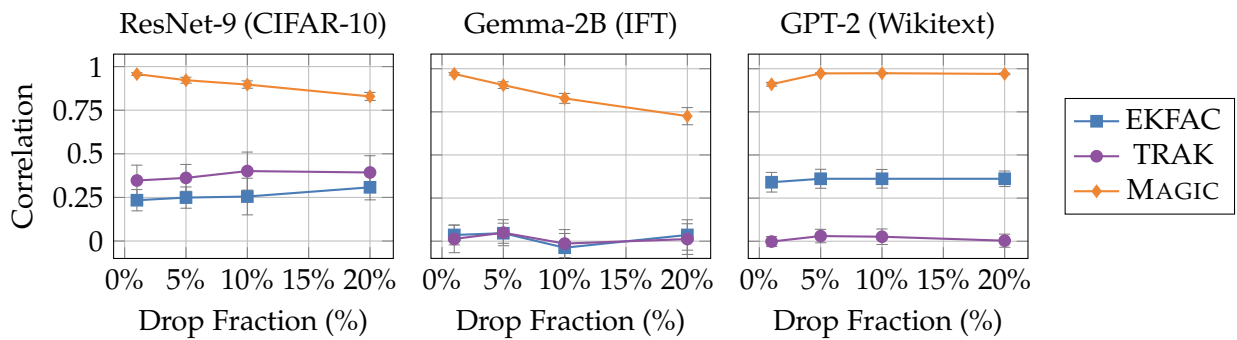
\begin{figure}[bp!]
  \centering
  \usetikzlibrary{positioning}
\usetikzlibrary{matrix}
\pgfplotsset{colormap/Set1}

\definecolor{mycolor1}{RGB}{74,124,182}
\definecolor{mycolor2}{RGB}{142,82,159}
\definecolor{mycolor3}{RGB}{239,134,50}

\begin{tikzpicture}
  \begin{groupplot}[
    group style={group size=3 by 1, horizontal sep=0.5cm},
    height=4.3cm,
    width=5.3cm,
    xlabel={Drop Fraction (\%)},
    ylabel={Correlation},
    title style={yshift=-.15cm},
    ytick={0, 0.25, 0.5, 0.75, 1},
    ymin=-0.1,
    xtick={0,5,10,15,20},
    xticklabels={0\%,5\%,10\%,15\%,20\%},
    grid=both,
    cycle list={
      {mycolor1, mark=square*, thick, error bars/.cd, y dir=both, y explicit, error bar style={color=gray, line cap=round}},
      {mycolor2, mark=*, thick, error bars/.cd, y dir=both, y explicit, error bar style={color=gray, line cap=round}},
      {mycolor3, mark=diamond*, thick, error bars/.cd, y dir=both, y explicit, error bar style={color=gray, line cap=round}},
    },
    legend style={at={(3.4,0.5)}, anchor=west, legend columns=1},
  ]

  \nextgroupplot[title={ResNet-9 (CIFAR-10)}]
  \addplot+ %
  coordinates {
    (1, 0.234) +- (0.063, 0.061)
    (5, 0.249) +- (0.055, 0.061)
    (10, 0.255) +- (0.102, 0.105)
    (20, 0.308) +- (0.080, 0.072)
  };
  \addlegendentry{EKFAC}
  \addplot+ %
  coordinates {
    (1, 0.347) +- (0.098, 0.088)
    (5, 0.362) +- (0.078, 0.077)
    (10, 0.401) +- (0.110, 0.109)
    (20, 0.393) +- (0.109, 0.096)
  };
  \addlegendentry{TRAK}
  \addplot+ %
  coordinates {
    (1, 0.957) +- (0.008, 0.008)
    (5, 0.922) +- (0.017, 0.017)
    (10, 0.897) +- (0.022, 0.021)
    (20, 0.829) +- (0.021, 0.023)
  };
  \addlegendentry{\method}
  
  \nextgroupplot[title={Gemma-2B (IFT)}, ylabel={}, yticklabels=\empty]
  \addplot+ %
  coordinates {
    (1, 0.036) +- (0.0603, 0.0576)
    (5, 0.046) +- (0.0569, 0.0583)
    (10, -0.038) +- (0.0780, 0.0826)
    (20, 0.036) +- (0.0846, 0.0882)
  };
  \addplot+ %
  coordinates {
    (1, 0.013) +- (0.0716, 0.0798)
    (5, 0.049) +- (0.0745, 0.0754)
    (10, -0.014) +- (0.0799, 0.0812)
    (20, 0.012) +- (0.0856, 0.0890)
  };
  \addplot+ %
  coordinates {
    (1, 0.970) +- (0.0096, 0.0072)
    (5, 0.905) +- (0.0214, 0.0193)
    (10, 0.828) +- (0.0324, 0.0287)
    (20, 0.725) +- (0.0614, 0.0498)
  };

  \nextgroupplot[title={GPT-2 (Wikitext)}, ylabel={}, yticklabels=\empty]
  \addplot+ %
  coordinates {
    (1, 0.342) +- (0.055, 0.057)
    (5, 0.362) +- (0.056, 0.055)
    (10, 0.362) +- (0.051, 0.054)
    (20, 0.362) +- (0.046, 0.045)
  };
  \addplot+ %
  coordinates {
    (1, -0.002) +- (0.029, 0.029)
    (5, 0.030) +- (0.039, 0.039)
    (10, 0.026) +- (0.046, 0.045)
    (20, 0.003) +- (0.037, 0.038)
  };
  \addplot+ %
  coordinates {
    (1, 0.910) +- (0.011, 0.011)
    (5, 0.973) +- (0.003, 0.003)
    (10, 0.974) +- (0.003, 0.003)
    (20, 0.970) +- (0.005, 0.004)
  };
  
  \end{groupplot}
  
  \end{tikzpicture}
  \caption{Linear datamodeling score (LDS) vs. drop fraction across settings for
    \method{} and baselines. The estimates of \method{} consistently correlate
    with the true model outputs (LDS: near $1.0$ for small enough drop fraction)
    while baselines often do not (LDS: below $0.4$). LDS decreases with
    increasing drop fraction for \method{} (as the Taylor estimate moves further
  from the center).}
  \label{fig:cifar_lds}
\end{figure}

\subsection{Results}

As shown in Figure~\ref{fig:cifar_lds}, \method{} attains near-perfect
LDS across settings and drop-out fractions,
although our predictions slightly degrade as we drop more data.
Existing baselines are noisy in comparison; these methods' predictions only
weakly correlate with the ground truth model losses.

\begin{figure}[ht!]
  \newcommand{\dataset}{cifar}
  \newcommand{\exampleid}{0}
  \newcommand{\scatteropacity}{0.5}
  \newcommand{\setting}{ResNet-9 on CIFAR-10}
  \newcommand{\fracone}{1}
  \newcommand{\frachtwo}{5}
  \newcommand{\datacsvscatterone}{plots/scatter/scatter_results_cifar/ind_0_dropfrac_0.01.csv}
  \newcommand{\datacsvscattertwo}{plots/scatter/scatter_results_cifar/ind_0_dropfrac_0.05.csv}
  \pgfplotsset{colormap/Set1}

\definecolor{Set1Red}{RGB}{74,124,182}
\definecolor{Set1Blue}{RGB}{142,82,159}
\definecolor{Set1Green}{RGB}{239,134,50}

\usetikzlibrary{calc}

\def\datacsvcorrelations{plots/scatter/correlations_\dataset/dropfrac_\fracone/example_\exampleid.csv}

\pgfplotstableread[col sep=comma]{\datacsvcorrelations}\datatable

\pgfplotstablegetelem{0}{corr}\of\datatable
\pgfmathprintnumberto[fixed, precision=2]{\pgfplotsretval}{\trakcorrone}

\pgfplotstablegetelem{1}{corr}\of\datatable
\pgfmathprintnumberto[fixed, precision=2]{\pgfplotsretval}{\ekfaccorrone}

\pgfplotstablegetelem{2}{corr}\of\datatable
\pgfmathprintnumberto[fixed, precision=2]{\pgfplotsretval}{\methodcorrone}

\def\dataset{wikitext}
\def\datacsvcorrelations{plots/scatter/correlations_\dataset/dropfrac_\frachtwo/example_\exampleid.csv}

\pgfplotstableread[col sep=comma]{\datacsvcorrelations}\datatable

\pgfplotstablegetelem{0}{corr}\of\datatable
\pgfmathprintnumberto[fixed, precision=2]{\pgfplotsretval}{\trakcorrhtwo}

\pgfplotstablegetelem{1}{corr}\of\datatable
\pgfmathprintnumberto[fixed, precision=2]{\pgfplotsretval}{\ekfaccorrhtwo}

\pgfplotstablegetelem{2}{corr}\of\datatable
\pgfmathprintnumberto[fixed, precision=2]{\pgfplotsretval}{\methodcorrhtwo}

\ifdefined\hidelegend
    \pgfplotsset{conditional legend style/.style={legend style={draw=none, fill=none}}}
\else
    \pgfplotsset{conditional legend style/.style={
        legend style={
            at={(0.12,1.22)}, %
            anchor=south, %
            legend columns=4, %
            /tikz/every even column/.append style={column sep=0.5cm}, %
            inner xsep=24pt, %
            inner ysep=2.5pt,  %
            legend image post style={scale=2.5} %
        }
    }}
\fi

\begin{tikzpicture}
    \node (settingbox) [minimum width=0.33\textwidth, 
                        minimum height=0.2\textwidth, 
                        anchor=north west, 
                        align=center] {\setting \\[.5em]
                        Correlation {\bf on Test Example \#$\mathbf{\exampleid}$} \\[.5em]
                        
        \begin{tabular}{ccc}
            \toprule
            Method & $\rho\ (\fracone\%)$ & $\rho\ (\frachtwo\%)$ \\
            \midrule
            EK-FAC & $\ekfaccorrone$ & $\ekfaccorrhtwo$ \\
            TRAK & $\trakcorrone$ & $\trakcorrhtwo$ \\
            \method & $\mathbf{\methodcorrone}$ & $\mathbf{\methodcorrhtwo}$ \\
            \bottomrule
        \end{tabular}
    };

    \begin{axis}[
        name=plot1,
        at={($(settingbox.east)+(2cm,0)$)}, %
        anchor=west, %
        width=0.33\textwidth,
        height=0.33\textwidth,
        xlabel={Predicted Loss},
        ylabel={True Loss},
        title={Drop \fracone\%},
        title style={yshift=-.15cm},
        conditional legend style,
        grid=both,
        grid style={line width=.1pt, draw=gray!10},
        major grid style={line width=.2pt,draw=gray!50},
        scaled y ticks=true,
        yticklabel style={/pgf/number format/.cd, fixed, precision=3},
        xticklabel style={/pgf/number format/.cd, fixed, precision=3},
        after end axis/.code={
            \pgfmathsetmacro{\myxmin}{\pgfkeysvalueof{/pgfplots/xmin}}
            \pgfmathsetmacro{\myxmax}{\pgfkeysvalueof{/pgfplots/xmax}}
            \pgfmathsetmacro{\myymin}{\pgfkeysvalueof{/pgfplots/ymin}}
            \pgfmathsetmacro{\myymax}{\pgfkeysvalueof{/pgfplots/ymax}}
            \pgfmathsetmacro{\linestart}{max(\myxmin, \myymin)}
            \pgfmathsetmacro{\lineend}{min(\myxmax, \myymax)}
            \draw[dashed, gray, very thick, opacity=0.5] 
                (axis cs:\linestart,\linestart) -- 
                (axis cs:\lineend,\lineend);
            }
    ]
        \addplot[
            only marks,
            mark=square*,
            mark size=1.5pt,
            color=Set1Red,
            fill opacity=\scatteropacity,
            draw opacity=\scatteropacity,
            point meta=explicit,
        ] table[x=kron_infl_10, y=true, meta expr=1, col sep=comma] {\datacsvscatterone};
        \ifdefined\hidelegend\else
            \addlegendentry{EK-FAC (re-scaled)}
        \fi

        \addplot[
            only marks,
            mark=triangle*,
            mark size=1.5pt,
            color=Set1Blue,
            fill opacity=\scatteropacity,
            draw opacity=\scatteropacity,
            point meta=explicit,
        ] table[x=trak_infl_10, y=true, meta expr=2, col sep=comma] {\datacsvscatterone};
        \ifdefined\hidelegend\else
            \addlegendentry{TRAK (re-scaled)}
        \fi

        \addplot[
            only marks,
            mark=*,
            mark size=2pt,
            color=Set1Green,
            fill opacity=1,
            draw opacity=1,
            point meta=explicit,
        ] table[x=infls, y=true, meta expr=0, col sep=comma] {\datacsvscatterone};
        \ifdefined\hidelegend\else
            \addlegendentry{\method{} (unscaled)}
        \fi

        \ifdefined\hidelegend\else
            \addlegendimage{dashed, gray, very thick,
                legend image code/.code={
                    \draw[dashed, gray, very thick] (0cm,0cm) -- (0.32cm,0cm); %
                }
            }
            \addlegendentry{\ $y=x$}
        \fi
    \end{axis}

    \begin{axis}[
        name=plot2,
        at={($(plot1.east)+(1.2cm,0)$)},
        anchor=west,
        width=0.33\textwidth,
        height=0.33\textwidth,
        xlabel={Predicted Loss},
        ylabel={},
        title={Drop \frachtwo\%},
        title style={yshift=-.15cm},
        grid=both,
        grid style={line width=.1pt, draw=gray!10},
        major grid style={line width=.2pt,draw=gray!50},
        scaled y ticks=true,
        yticklabel style={/pgf/number format/.cd, fixed, precision=3},
        xticklabel style={/pgf/number format/.cd, fixed, precision=3},
        after end axis/.code={
            \pgfmathsetmacro{\myxmin}{\pgfkeysvalueof{/pgfplots/xmin}}
            \pgfmathsetmacro{\myxmax}{\pgfkeysvalueof{/pgfplots/xmax}}
            \pgfmathsetmacro{\myymin}{\pgfkeysvalueof{/pgfplots/ymin}}
            \pgfmathsetmacro{\myymax}{\pgfkeysvalueof{/pgfplots/ymax}}
            \pgfmathsetmacro{\linestart}{max(\myxmin, \myymin)}
            \pgfmathsetmacro{\lineend}{min(\myxmax, \myymax)}
            \draw[dashed, gray, very thick, opacity=0.5] 
                (axis cs:\linestart,\linestart) -- 
                (axis cs:\lineend,\lineend);
            }
    ]
        \addplot[
            only marks,
            mark=square*,
            mark size=1.5pt,
            color=Set1Red,
            fill opacity=\scatteropacity,
            draw opacity=\scatteropacity,
            point meta=explicit,
        ] table[x=kron_infl_10, y=true, meta expr=1, col sep=comma] {\datacsvscattertwo};
        
        \addplot[
            only marks,
            mark=triangle*,
            mark size=1.5pt,
            color=Set1Blue,
            fill opacity=\scatteropacity,
            draw opacity=\scatteropacity,
            point meta=explicit,
        ] table[x=trak_infl_10, y=true, meta expr=2, col sep=comma] {\datacsvscattertwo};
        
        \addplot[
            only marks,
            mark=*,
            mark size=2pt,
            color=Set1Green,
            fill opacity=1,
            draw opacity=1,
            point meta=explicit,
        ] table[x=infls, y=true, meta expr=0, col sep=comma] {\datacsvscattertwo};
    \end{axis}
    
\end{tikzpicture}

  \renewcommand{\dataset}{gemma}
  \renewcommand{\exampleid}{17}
  \renewcommand{\scatteropacity}{0.5}
  \renewcommand{\setting}{Gemma-2B on IFT data}
  \newcommand{\hidelegend}{true}
  \renewcommand{\fracone}{1}
  \renewcommand{\frachtwo}{5}
  \renewcommand{\datacsvscatterone}{plots/scatter/scatter_results_gemma/example_17_dropfrac_01.csv}
  \renewcommand{\datacsvscattertwo}{plots/scatter/scatter_results_gemma/example_17_dropfrac_05.csv}
  \pgfplotsset{colormap/Set1}

\definecolor{Set1Red}{RGB}{74,124,182}
\definecolor{Set1Blue}{RGB}{142,82,159}
\definecolor{Set1Green}{RGB}{239,134,50}

\usetikzlibrary{calc}

\def\datacsvcorrelations{plots/scatter/correlations_\dataset/dropfrac_\fracone/example_\exampleid.csv}

\pgfplotstableread[col sep=comma]{\datacsvcorrelations}\datatable

\pgfplotstablegetelem{0}{corr}\of\datatable
\pgfmathprintnumberto[fixed, precision=2]{\pgfplotsretval}{\trakcorrone}

\pgfplotstablegetelem{1}{corr}\of\datatable
\pgfmathprintnumberto[fixed, precision=2]{\pgfplotsretval}{\ekfaccorrone}

\pgfplotstablegetelem{2}{corr}\of\datatable
\pgfmathprintnumberto[fixed, precision=2]{\pgfplotsretval}{\methodcorrone}

\def\dataset{wikitext}
\def\datacsvcorrelations{plots/scatter/correlations_\dataset/dropfrac_\frachtwo/example_\exampleid.csv}

\pgfplotstableread[col sep=comma]{\datacsvcorrelations}\datatable

\pgfplotstablegetelem{0}{corr}\of\datatable
\pgfmathprintnumberto[fixed, precision=2]{\pgfplotsretval}{\trakcorrhtwo}

\pgfplotstablegetelem{1}{corr}\of\datatable
\pgfmathprintnumberto[fixed, precision=2]{\pgfplotsretval}{\ekfaccorrhtwo}

\pgfplotstablegetelem{2}{corr}\of\datatable
\pgfmathprintnumberto[fixed, precision=2]{\pgfplotsretval}{\methodcorrhtwo}

\ifdefined\hidelegend
    \pgfplotsset{conditional legend style/.style={legend style={draw=none, fill=none}}}
\else
    \pgfplotsset{conditional legend style/.style={
        legend style={
            at={(0.12,1.22)}, %
            anchor=south, %
            legend columns=4, %
            /tikz/every even column/.append style={column sep=0.5cm}, %
            inner xsep=24pt, %
            inner ysep=2.5pt,  %
            legend image post style={scale=2.5} %
        }
    }}
\fi

\begin{tikzpicture}
    \node (settingbox) [minimum width=0.33\textwidth, 
                        minimum height=0.2\textwidth, 
                        anchor=north west, 
                        align=center] {\setting \\[.5em]
                        Correlation {\bf on Test Example \#$\mathbf{\exampleid}$} \\[.5em]
                        
        \begin{tabular}{ccc}
            \toprule
            Method & $\rho\ (\fracone\%)$ & $\rho\ (\frachtwo\%)$ \\
            \midrule
            EK-FAC & $\ekfaccorrone$ & $\ekfaccorrhtwo$ \\
            TRAK & $\trakcorrone$ & $\trakcorrhtwo$ \\
            \method & $\mathbf{\methodcorrone}$ & $\mathbf{\methodcorrhtwo}$ \\
            \bottomrule
        \end{tabular}
    };

    \begin{axis}[
        name=plot1,
        at={($(settingbox.east)+(2cm,0)$)}, %
        anchor=west, %
        width=0.33\textwidth,
        height=0.33\textwidth,
        xlabel={Predicted Loss},
        ylabel={True Loss},
        title={Drop \fracone\%},
        title style={yshift=-.15cm},
        conditional legend style,
        grid=both,
        grid style={line width=.1pt, draw=gray!10},
        major grid style={line width=.2pt,draw=gray!50},
        scaled y ticks=true,
        yticklabel style={/pgf/number format/.cd, fixed, precision=3},
        xticklabel style={/pgf/number format/.cd, fixed, precision=3},
        after end axis/.code={
            \pgfmathsetmacro{\myxmin}{\pgfkeysvalueof{/pgfplots/xmin}}
            \pgfmathsetmacro{\myxmax}{\pgfkeysvalueof{/pgfplots/xmax}}
            \pgfmathsetmacro{\myymin}{\pgfkeysvalueof{/pgfplots/ymin}}
            \pgfmathsetmacro{\myymax}{\pgfkeysvalueof{/pgfplots/ymax}}
            \pgfmathsetmacro{\linestart}{max(\myxmin, \myymin)}
            \pgfmathsetmacro{\lineend}{min(\myxmax, \myymax)}
            \draw[dashed, gray, very thick, opacity=0.5] 
                (axis cs:\linestart,\linestart) -- 
                (axis cs:\lineend,\lineend);
            }
    ]
        \addplot[
            only marks,
            mark=square*,
            mark size=1.5pt,
            color=Set1Red,
            fill opacity=\scatteropacity,
            draw opacity=\scatteropacity,
            point meta=explicit,
        ] table[x=kron_infl_10, y=true, meta expr=1, col sep=comma] {\datacsvscatterone};
        \ifdefined\hidelegend\else
            \addlegendentry{EK-FAC (re-scaled)}
        \fi

        \addplot[
            only marks,
            mark=triangle*,
            mark size=1.5pt,
            color=Set1Blue,
            fill opacity=\scatteropacity,
            draw opacity=\scatteropacity,
            point meta=explicit,
        ] table[x=trak_infl_10, y=true, meta expr=2, col sep=comma] {\datacsvscatterone};
        \ifdefined\hidelegend\else
            \addlegendentry{TRAK (re-scaled)}
        \fi

        \addplot[
            only marks,
            mark=*,
            mark size=2pt,
            color=Set1Green,
            fill opacity=1,
            draw opacity=1,
            point meta=explicit,
        ] table[x=infls, y=true, meta expr=0, col sep=comma] {\datacsvscatterone};
        \ifdefined\hidelegend\else
            \addlegendentry{\method{} (unscaled)}
        \fi

        \ifdefined\hidelegend\else
            \addlegendimage{dashed, gray, very thick,
                legend image code/.code={
                    \draw[dashed, gray, very thick] (0cm,0cm) -- (0.32cm,0cm); %
                }
            }
            \addlegendentry{\ $y=x$}
        \fi
    \end{axis}

    \begin{axis}[
        name=plot2,
        at={($(plot1.east)+(1.2cm,0)$)},
        anchor=west,
        width=0.33\textwidth,
        height=0.33\textwidth,
        xlabel={Predicted Loss},
        ylabel={},
        title={Drop \frachtwo\%},
        title style={yshift=-.15cm},
        grid=both,
        grid style={line width=.1pt, draw=gray!10},
        major grid style={line width=.2pt,draw=gray!50},
        scaled y ticks=true,
        yticklabel style={/pgf/number format/.cd, fixed, precision=3},
        xticklabel style={/pgf/number format/.cd, fixed, precision=3},
        after end axis/.code={
            \pgfmathsetmacro{\myxmin}{\pgfkeysvalueof{/pgfplots/xmin}}
            \pgfmathsetmacro{\myxmax}{\pgfkeysvalueof{/pgfplots/xmax}}
            \pgfmathsetmacro{\myymin}{\pgfkeysvalueof{/pgfplots/ymin}}
            \pgfmathsetmacro{\myymax}{\pgfkeysvalueof{/pgfplots/ymax}}
            \pgfmathsetmacro{\linestart}{max(\myxmin, \myymin)}
            \pgfmathsetmacro{\lineend}{min(\myxmax, \myymax)}
            \draw[dashed, gray, very thick, opacity=0.5] 
                (axis cs:\linestart,\linestart) -- 
                (axis cs:\lineend,\lineend);
            }
    ]
        \addplot[
            only marks,
            mark=square*,
            mark size=1.5pt,
            color=Set1Red,
            fill opacity=\scatteropacity,
            draw opacity=\scatteropacity,
            point meta=explicit,
        ] table[x=kron_infl_10, y=true, meta expr=1, col sep=comma] {\datacsvscattertwo};
        
        \addplot[
            only marks,
            mark=triangle*,
            mark size=1.5pt,
            color=Set1Blue,
            fill opacity=\scatteropacity,
            draw opacity=\scatteropacity,
            point meta=explicit,
        ] table[x=trak_infl_10, y=true, meta expr=2, col sep=comma] {\datacsvscattertwo};
        
        \addplot[
            only marks,
            mark=*,
            mark size=2pt,
            color=Set1Green,
            fill opacity=1,
            draw opacity=1,
            point meta=explicit,
        ] table[x=infls, y=true, meta expr=0, col sep=comma] {\datacsvscattertwo};
    \end{axis}
    
\end{tikzpicture}

  \renewcommand{\dataset}{wikitext}
  \renewcommand{\exampleid}{32}
  \renewcommand{\scatteropacity}{0.5}
  \renewcommand{\setting}{GPT-2 on WikiText}
  \renewcommand{\hidelegend}{true}
  \renewcommand{\fracone}{1}
  \renewcommand{\frachtwo}{5}
  \renewcommand{\datacsvscatterone}{plots/scatter/scatter_results_wikitext/example_32_dropfrac_0.01.csv}
  \renewcommand{\datacsvscattertwo}{plots/scatter/scatter_results_wikitext/example_32_dropfrac_0.05.csv}
  \pgfplotsset{colormap/Set1}

\definecolor{Set1Red}{RGB}{74,124,182}
\definecolor{Set1Blue}{RGB}{142,82,159}
\definecolor{Set1Green}{RGB}{239,134,50}

\usetikzlibrary{calc}

\def\datacsvcorrelations{plots/scatter/correlations_\dataset/dropfrac_\fracone/example_\exampleid.csv}

\pgfplotstableread[col sep=comma]{\datacsvcorrelations}\datatable

\pgfplotstablegetelem{0}{corr}\of\datatable
\pgfmathprintnumberto[fixed, precision=2]{\pgfplotsretval}{\trakcorrone}

\pgfplotstablegetelem{1}{corr}\of\datatable
\pgfmathprintnumberto[fixed, precision=2]{\pgfplotsretval}{\ekfaccorrone}

\pgfplotstablegetelem{2}{corr}\of\datatable
\pgfmathprintnumberto[fixed, precision=2]{\pgfplotsretval}{\methodcorrone}

\def\dataset{wikitext}
\def\datacsvcorrelations{plots/scatter/correlations_\dataset/dropfrac_\frachtwo/example_\exampleid.csv}

\pgfplotstableread[col sep=comma]{\datacsvcorrelations}\datatable

\pgfplotstablegetelem{0}{corr}\of\datatable
\pgfmathprintnumberto[fixed, precision=2]{\pgfplotsretval}{\trakcorrhtwo}

\pgfplotstablegetelem{1}{corr}\of\datatable
\pgfmathprintnumberto[fixed, precision=2]{\pgfplotsretval}{\ekfaccorrhtwo}

\pgfplotstablegetelem{2}{corr}\of\datatable
\pgfmathprintnumberto[fixed, precision=2]{\pgfplotsretval}{\methodcorrhtwo}

\ifdefined\hidelegend
    \pgfplotsset{conditional legend style/.style={legend style={draw=none, fill=none}}}
\else
    \pgfplotsset{conditional legend style/.style={
        legend style={
            at={(0.12,1.22)}, %
            anchor=south, %
            legend columns=4, %
            /tikz/every even column/.append style={column sep=0.5cm}, %
            inner xsep=24pt, %
            inner ysep=2.5pt,  %
            legend image post style={scale=2.5} %
        }
    }}
\fi

\begin{tikzpicture}
    \node (settingbox) [minimum width=0.33\textwidth, 
                        minimum height=0.2\textwidth, 
                        anchor=north west, 
                        align=center] {\setting \\[.5em]
                        Correlation {\bf on Test Example \#$\mathbf{\exampleid}$} \\[.5em]
                        
        \begin{tabular}{ccc}
            \toprule
            Method & $\rho\ (\fracone\%)$ & $\rho\ (\frachtwo\%)$ \\
            \midrule
            EK-FAC & $\ekfaccorrone$ & $\ekfaccorrhtwo$ \\
            TRAK & $\trakcorrone$ & $\trakcorrhtwo$ \\
            \method & $\mathbf{\methodcorrone}$ & $\mathbf{\methodcorrhtwo}$ \\
            \bottomrule
        \end{tabular}
    };

    \begin{axis}[
        name=plot1,
        at={($(settingbox.east)+(2cm,0)$)}, %
        anchor=west, %
        width=0.33\textwidth,
        height=0.33\textwidth,
        xlabel={Predicted Loss},
        ylabel={True Loss},
        title={Drop \fracone\%},
        title style={yshift=-.15cm},
        conditional legend style,
        grid=both,
        grid style={line width=.1pt, draw=gray!10},
        major grid style={line width=.2pt,draw=gray!50},
        scaled y ticks=true,
        yticklabel style={/pgf/number format/.cd, fixed, precision=3},
        xticklabel style={/pgf/number format/.cd, fixed, precision=3},
        after end axis/.code={
            \pgfmathsetmacro{\myxmin}{\pgfkeysvalueof{/pgfplots/xmin}}
            \pgfmathsetmacro{\myxmax}{\pgfkeysvalueof{/pgfplots/xmax}}
            \pgfmathsetmacro{\myymin}{\pgfkeysvalueof{/pgfplots/ymin}}
            \pgfmathsetmacro{\myymax}{\pgfkeysvalueof{/pgfplots/ymax}}
            \pgfmathsetmacro{\linestart}{max(\myxmin, \myymin)}
            \pgfmathsetmacro{\lineend}{min(\myxmax, \myymax)}
            \draw[dashed, gray, very thick, opacity=0.5] 
                (axis cs:\linestart,\linestart) -- 
                (axis cs:\lineend,\lineend);
            }
    ]
        \addplot[
            only marks,
            mark=square*,
            mark size=1.5pt,
            color=Set1Red,
            fill opacity=\scatteropacity,
            draw opacity=\scatteropacity,
            point meta=explicit,
        ] table[x=kron_infl_10, y=true, meta expr=1, col sep=comma] {\datacsvscatterone};
        \ifdefined\hidelegend\else
            \addlegendentry{EK-FAC (re-scaled)}
        \fi

        \addplot[
            only marks,
            mark=triangle*,
            mark size=1.5pt,
            color=Set1Blue,
            fill opacity=\scatteropacity,
            draw opacity=\scatteropacity,
            point meta=explicit,
        ] table[x=trak_infl_10, y=true, meta expr=2, col sep=comma] {\datacsvscatterone};
        \ifdefined\hidelegend\else
            \addlegendentry{TRAK (re-scaled)}
        \fi

        \addplot[
            only marks,
            mark=*,
            mark size=2pt,
            color=Set1Green,
            fill opacity=1,
            draw opacity=1,
            point meta=explicit,
        ] table[x=infls, y=true, meta expr=0, col sep=comma] {\datacsvscatterone};
        \ifdefined\hidelegend\else
            \addlegendentry{\method{} (unscaled)}
        \fi

        \ifdefined\hidelegend\else
            \addlegendimage{dashed, gray, very thick,
                legend image code/.code={
                    \draw[dashed, gray, very thick] (0cm,0cm) -- (0.32cm,0cm); %
                }
            }
            \addlegendentry{\ $y=x$}
        \fi
    \end{axis}

    \begin{axis}[
        name=plot2,
        at={($(plot1.east)+(1.2cm,0)$)},
        anchor=west,
        width=0.33\textwidth,
        height=0.33\textwidth,
        xlabel={Predicted Loss},
        ylabel={},
        title={Drop \frachtwo\%},
        title style={yshift=-.15cm},
        grid=both,
        grid style={line width=.1pt, draw=gray!10},
        major grid style={line width=.2pt,draw=gray!50},
        scaled y ticks=true,
        yticklabel style={/pgf/number format/.cd, fixed, precision=3},
        xticklabel style={/pgf/number format/.cd, fixed, precision=3},
        after end axis/.code={
            \pgfmathsetmacro{\myxmin}{\pgfkeysvalueof{/pgfplots/xmin}}
            \pgfmathsetmacro{\myxmax}{\pgfkeysvalueof{/pgfplots/xmax}}
            \pgfmathsetmacro{\myymin}{\pgfkeysvalueof{/pgfplots/ymin}}
            \pgfmathsetmacro{\myymax}{\pgfkeysvalueof{/pgfplots/ymax}}
            \pgfmathsetmacro{\linestart}{max(\myxmin, \myymin)}
            \pgfmathsetmacro{\lineend}{min(\myxmax, \myymax)}
            \draw[dashed, gray, very thick, opacity=0.5] 
                (axis cs:\linestart,\linestart) -- 
                (axis cs:\lineend,\lineend);
            }
    ]
        \addplot[
            only marks,
            mark=square*,
            mark size=1.5pt,
            color=Set1Red,
            fill opacity=\scatteropacity,
            draw opacity=\scatteropacity,
            point meta=explicit,
        ] table[x=kron_infl_10, y=true, meta expr=1, col sep=comma] {\datacsvscattertwo};
        
        \addplot[
            only marks,
            mark=triangle*,
            mark size=1.5pt,
            color=Set1Blue,
            fill opacity=\scatteropacity,
            draw opacity=\scatteropacity,
            point meta=explicit,
        ] table[x=trak_infl_10, y=true, meta expr=2, col sep=comma] {\datacsvscattertwo};
        
        \addplot[
            only marks,
            mark=*,
            mark size=2pt,
            color=Set1Green,
            fill opacity=1,
            draw opacity=1,
            point meta=explicit,
        ] table[x=infls, y=true, meta expr=0, col sep=comma] {\datacsvscattertwo};
    \end{axis}
    
\end{tikzpicture}

  \caption{Results of \method{} and baselines on randomly chosen, individual
    samples from the three settings we consider: CIFAR-10, Gemma-2B, and GPT-2.
    We evaluate by predicting model output after dropping a random 1\%/5\% of
    the data (cf. \eqref{eq:lds}) and plotting the results against the true
    model output for that drop set. \method{} estimates consistently highly
    correlate with the true output across settings (for small enough training
  data drop fractions).}
  \label{fig:scatters}
\end{figure}
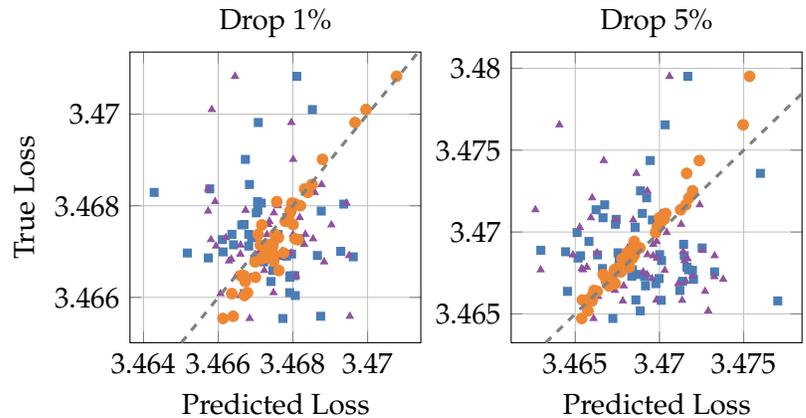

In Figure~\ref{fig:scatters}, we randomly select a test example 
from each scenario, and plot predictions of test loss against true test loss for
each data attribution method. \method{} almost exactly predicts the true test
loss, even in absolute terms. On the other hand, baselines barely correlate
with the true predictions, and are mis-scaled in absolute terms
(TRAK and EKFAC predictions are not of the right order of magnitude,
so we rescale them to visualize them on the same plot).

\medskip
\noindent \textbf{Optimality of \method{}.} We observe that the performance of
\method{} degrades with the fraction of samples dropped. While \method{} has
near perfect LDS when predicting the effect of removing a small fraction of the
training data (i.e., 1\%), the LDS degrades at larger drop-out fractions
(i.e., 20\%). Our near-perfect performance when dropping only a few points
indicates that our method is in some sense optimal among linear predictors: 
we can perfectly predict in a small ball around ``not dropping out any points,'' 
but curvature (in training data weight space) 
causes the linear approximation to degrade
further away from training on all the data.

\section{Discussion}
\label{sec:discussion}
In this section, we discuss the 
connections between our single-model data attribution setting and 
the standard data attribution problem, 
the computational cost of our \method{} compared to the baselines 
we consider, and finally some limitations of our method.

\paragraph{Relationship between single-model and standard data attribution.}
Recall (from Section \ref{sec:single_model_data_attribution}) that single-model data attribution 
differs from the standard task of predictive data
attribution in that there is no randomness. From a conceptual perspective, 
this means that while standard data attribution is about predicting how a 
{\em new} model would behave if retrained from scratch on a counterfactual
dataset, single-model data attribution is about predicting how a {\em specific}
model would behave if the data had been different.

Technically, single-model data attribution is a more difficult problem than
standard data attribution in the following natural sense: a {\em perfect}
single-model data attribution method can exactly predict the behavior 
of a specific model on any counterfactual dataset. 
By averaging these predictions over multiple specific models 
(each corresponding to a different learning algorithm seed),
we obtain a perfect ``standard'' predictive data attribution method.
The other direction, however, does not hold: it may be possible to construct
a predictive data attribution method that perfectly predicts the 
average behavior of a learning algorithm, but not the behavior
of a specific model.

\paragraph{Computational cost of data attribution.}
One consideration that we have not yet addressed is the computational cost
of {\em building} the data attribution method $\hat{f}$, which varies
per-method and can differ greatly asymptotically for different scenarios.
In what follows, we compare the cost of building \method{},
TRAK, and EK-FAC in a setting with $N$ training samples and $n$ test samples.
\begin{itemize}
  \item TRAK \citep{park2023trak} has a hyperparameter $k$ which 
  controls the accuracy of the method (mechanistically, $k$ 
  determines the number of intermediate checkpoints used by the method).
  TRAK with $k$ averaged models requires computing
  $k \cdot (N + n)$ per-sample gradients, and storing small 
  randomly-projected versions of them. 
  \item EK-FAC has a similar computational structure to TRAK, but 
  instead of just randomly projecting the per-sample gradients, 
  it applies an iterative algorithm with the goal of better 
  approximating the Hessian. 
  In practice, this algorithm seems to dominate 
  its runtime---we refer the reader to \citep{george2018fast} for more details.
  \item \method{} requires a forward and backward pass through the model
    training process for each of the $n$ test samples. The cost 
    of the forward and backward pass scale linearly 
    in the number of train samples $N$, 
    so the full complexity is on the order of $N\cdot n$.
\end{itemize}
In summary, \method{} is faster than TRAK and EK-FAC when $n$ is
small---for example, when $n = 1$ (attributing a single test sample)
on the CIFAR-10 dataset, 
\method{} costs essentially 3-5x as much as training a single model
(15 minutes on a single A100 GPU),
while TRAK and EK-FAC are both 20-100x slower.
As $n$ increases, however, \method{}'s cost increases linearly,
while TRAK and EK-FAC's costs stay roughly constant.

\paragraph{Limitations and future work.}
One prominent limitation is the one discussed in the previous paragraph: 
\method{} becomes expensive as the number of test samples $n$ increases.
The reason for this is that each test sample requires its 
own metagradient calculation, and each of these calculations 
costs 3-5\texttimes{} as much as model training in practice
(see \citet{engstrom2025optimizing} for an in-depth discussion).
An interesting avenue for future work is to develop (a) more efficient 
methods for calculating metagradients, and (b) ways to leverage metagradients to
compute data attribution for multiple test samples at once.

On the algorithmic side, \method{} can only attribute for learning algorithms
that are smooth 
(i.e., as described by \citet{engstrom2025optimizing}, 
``metasmooth''). This and
previous work has identified variations on standard learning algorithms that
are smooth---including for language models,
CLIP~\citep{radford2021learning} models, and standard vision models---but
finding such a variation can take work (see Section 2 of
\citet{engstrom2025optimizing} for a discussion). 
As a simple workaround, we have found
that simply pretraining a model state for a few hundred iterates
(and treating the resulting weights as a ``pretrained'' model initialization) 
is enough to induce smoothness.
Depending on the scenario, this workaround may or may not be acceptable
(i.e., using this method will not allow one to account for the impact of
{\em all} data, but only the data seen after the ``initialization'' stage).

\section{Related Work}
Our work is related to the growing literature on predictive data attribution
methods---see \citet{ilyas2024data} or \citet{hammoudeh2022training} for 
a survey. 
Related to our method are those that use variants of the influence
function~\citep{hampel2011robust} (i.e., the Taylor 
approximation from Section \ref{sec:influence_fn}). 
In settings where the learning algorithm 
minimizes a convex objective, such influence function-based methods 
are known to have an efficient and accurate closed form 
\citep{rad2018scalable,koh2019accuracy,giordano2019swiss,nobel2024randalo}.
In scenarios where the learning algorithm does not return
a convex minimizer (such as in deep learning), 
this closed form is not available.

In such cases, the dominant approach is to apply one of many
efficient approximate approaches
\citep{koh2017understanding,ladhak2022tracing,schioppa2022scaling,park2023trak,grosse2023studying}.
However, in the non-convex settting, these approximations 
do not offer correctness guarantees like they do in convex settings,
and can even have different interpretations entirely \citep{bae2022if}---potentially 
leading to the poor correlations we observe in Section~\ref{sec:results}.
Closer to our work are methods that attempt to approximate the influence function 
via {\em unrolling} \citep{bae2024training}. These methods leverage 
the same recursive formula of \eqref{eq:replay_meta_grad}---but 
still {\em approximate} the influence function rather than compute it exactly.

To compute the influence function exactly,
we leverage recent advances in {\em metagradient} calculation 
\citep{engstrom2025optimizing}. These advances in turn build on 
a long line of work on differentiating through optimization 
\citep{maclaurin2015gradient,lorraine2020optimizing}---see 
\citet{engstrom2025optimizing} for a recent survey.

Finally, our single-model data attribution setting is motivated by the 
nondeterminism of model training. 
This phenomenon has been studied from a variety of perspectives, 
including training dynamics~\citep{zhuang2022randomness,jordan2024variance}, 
fairness~\citep{black2022model,marx2020predictive}, 
and even data attribution~\citep{ilyas2022datamodels,nguyen2023bayesian}.

\section{Conclusion}
We present \method{}, a new data attribution method that near-exactly predicts
how model outputs change as a function of its training data (according to
standard metrics). To do so, \method{} operates by calculating the
exact influence function using recent advances in
metadifferentiation~\citep{engstrom2025optimizing}. Given the magnitude at which
\method{} improves our ability to estimate the effect of training data at high
fidelity, we are excited to see what downstream applications  for data
attribution \method{} unlocks, including near-perfect unlearning
\citep{georgiev2024attribute}, model debugging
\citep{koh2017understanding,shah2022modeldiff}, and more.

\section{Acknowledgments}
Work supported in part by the NSF grant DMS-2134108 and Open Philanthropy, and
in part by NSF Grant No. 2346519. The authors would like to thank Benjamin
Chen, Aleksander M\k{a}dry, Axel Feldmann, Billy Moses, Joel Flynn, Sam Park,
Sarah Cen, Sam Hopkins, and Piotr Indyk for helpful references as well as
discussions and feedback on early versions of this work.

\printbibliography

\clearpage
\appendix

\section{Experimental details}
\label{app:expdetails}
In this section, we provide additional details on the experimental setup used in
the main paper, including the training details of the models,
and the datasets used.

\paragraph{ResNet-9 on CIFAR-10.} We use the ResNet-9 architecture from
\citep{jordan202494}, with the hyperparameters given in
Table~\ref{tab:resnet9_cifar10_hyperparams}. To give concrete details: the
training set $S$ comprises $50,000$ CIFAR-10 training samples, the
learning algorithm
$\mathcal{A}$ is standard supervised training, and we consider 50 measurement
functions $\phi_i$ corresponding to loss on 50 different CIFAR-10 test samples.

\begin{table}[h]
  \centering
  \begin{tabular}{ll}
    \toprule
    Hyperparameter & Value \\
    \midrule
    Learning rate & 1.2 \\
    Weight decay & 0.001 \\
    Bias scale & 8.0 \\
    Batch size & 1000 \\
    Epochs & 12 \\
    Final layer scale & 0.04 \\
    Momentum & 0.875 \\
    Pooling type & Log-sum-exp \\
    Pooling $\varepsilon$ & 0.1 \\
    Width multiplier & 2.5 \\
    LR schedule & One-cycle Linear \\
    LR start multiplier & 0.07 \\
    LR end multiplier & 0.2 \\
    LR peak time & 0.5 \\
    \bottomrule
  \end{tabular}
  \caption{Hyperparameters for ResNet-9 on CIFAR-10.}
  \label{tab:resnet9_cifar10_hyperparams}
\end{table}

\paragraph{Gemma-2B LoRA on IFT Data.} We use the variant of
LESS~\citep{xia2024less} from \citet{engstrom2025optimizing}. In particular, the
training dataset consists of the four instruction fine-tuning sets seen in
Table~\ref{tab:ift_training_dataset} as in LESS. The total number of points is
around $300,000$ and is exactly four combined IFT datasets (Flan
  V2~\citep{longpre2023flan}, CoT~\citep{wei2022chain},
  DOLLY~\citep{conover2023free}, and Open Assistant 1
\citep{kopf2024openassistant}). We test on a randomly chosen (task balanced)
subset of of MMLU comprising 32 test samples. We use 4-shot in-context learning
for these samples. We adapt a LoRA to a Gemma-2B model (the pretraining-only
Gemma-2B model~\citep{team2024gemma}) using the LoRA configuration from
\citet{xia2024less}. For model training, we use the same setup as
\citet{engstrom2025optimizing}, except with $\epsilon_\textrm{root}=10^{-6}$. In
particular, we train with ADAM ($\beta_1=0.95, \beta_2=0.975$, decoupled weight
decay as $10^{-5}$) and a one-cycle linear schedule starting at $10^{-6}$ of the
maximum learning rate, reaching the peak over 25\% of training, then ending at
$0.1$ of the maximum learning rate ($0.0004$). We insert a positive
$\epsilon_\textrm{root}$ into the inverse square root term in the ADAM update to
prevent metagradient (and to a lesser extent update) blowup.

\paragraph{GPT-2 fine-tuning on Wikitext.} We optimize a pre-trained
GPT2~\citep{radford2019language} model on Wikitext~\citep{foundation2022english}
using causal language modeling. We split the Wikitext samples into size 512
context length chunks and into train and test splits, with 256 samples in the
test split and 4608 samples in the train split. We attribute on the test split,
and use 4 epochs of the train split during training. We use the same ADAM
optimizer setup above except that we set $\epsilon_\textrm{root}=10^{-8}$, max
learning rate to $0.0008$, and do not anneal ADAM $\epsilon_{\mathrm{root}}$.

\begin{table}[h]
  \small
  \centering
  \caption{Details of IFT training datasets.}
  \label{tab:ift_training_dataset}
  \begin{tabular}{l l l c c}
    \toprule
    Dataset & \# Instance & Sourced from & Prompt Len. & Complet. Len. \\
    \midrule
    FLAN V2 & 100,000 & NLP datasets and human-written instructions &
    355.7 & 31.2 \\
    CoT & 100,000 & NLP datasets and human-written CoTs & 266 & 53.2 \\
    Dolly & 15,011 & Human-written from scratch & 118.1 & 91.3 \\
    Open Ast. 1 & 55,668 & Human-written from scratch & 34.8 & 212.5 \\
    \bottomrule
  \end{tabular}
\end{table}

\section{Baselines}
\label{app:baselines}
In this section, we provide a brief overview of the baselines we consider.
The basis of both of these methods is similar, and rooted in the Taylor
approximation \eqref{eq:infl_fn} that also underpins \method{}:
\begin{equation*}
  \hat{f}(\wvec) := f(\mathbf{1}_n) +
  \left(\frac{\partial f(\wvec)}{\partial
  \wvec}\bigg|_{\wvec=\mathbf{1}_n}\right)^\top  (\wvec -
  \mathbf{1}_n);
\end{equation*}

\subsection{TRAK (Tracing the Randomly-projected After Kernel)}
TRAK \citep{park2023trak} estimates the influence of individual training
examples on model predictions. However, instead of computing exact the
influence, TRAK calculates the influence for a simple proxy model that (a) is
easy to calculate the influence for and (b) is meant to match the original model
class of interest. In particular, TRAK \textit{approximates} the original
learning algorithm, $\mathcal{A}$, by linearizing around the final parameters.
We refer the reader to \citet{park2023trak} for full details.

\subsection{EK-FAC (Eigenvalue-corrected Kronecker-Factored Approximate
Curvature)} EK-FAC \citep{bae2022if,grosse2023studying} is another
influence-based data attribution method. To estimate the influence function, the
method estimates the Hessian via the Fisher information matrix/Gauss-Newton
hessian~\citep{martens2015optimizing,george2018fast}, then applies a version of
the infinitesimal jackknife~\citep{giordano2019swiss} to calculate the gradient
with respect to data weights. For an excellent high level overview of this
approach, see Appendix D.1 of \citet{bae2024training}.

\end{document}